# Neural Style Transfer for Remote Sensing

M. Karatzoglidi [1] · G. Felekis [2] · E. Charou [3]

[1] National Technical University of Athens
[2] University College London
[3] NCSR Demokritos

**Abstract**

The well-known technique outlined in the paper of Leon A. Gatys et al., A Neural Algorithm of Artistic Style [1], has become a trending topic both in academic literature and industrial applications. Neural Style Transfer (NST) constitutes an essential tool for a wide range of applications, such as artistic stylization of 2D images, user-assisted creation tools and production tools for entertainment applications [2].
The purpose of this study is to present a method for creating artistic maps from satellite images, based on the NST algorithm. This method includes three basic steps (i) application of semantic image segmentation on the original satellite image, dividing its content into two classes (i.e. land, water), (ii) application of neural style transfer for each class and (iii) creation of a collage, i.e. an artistic image consisting of a combination of the two stylized image generated on the previous step.

**Keywords:** Neural style transfer, deep learning, satellite imagery, remote sensing,

1.  Introduction

Art has been around in one form or another, for thousands of years, including activities, ranging from painting and sculpting, to architecture and photography etc. Looking back in time, one can realise that the earliest form of art was painting, presented on ancient walls. Nowadays, with the development of computer systems, it is possible to create art driven by artificial intelligence (AI).
Not a long time ago, re-drawing an image in the style of a famous painting was a time-consuming task and required a well-trained artist. With the advancement of time, computer systems are becoming able to perform tasks that normally require human intelligence. Leon A. Gatys et al. [1] studied how to reproduce famous painting styles on natural images, taking advantage of the capabilities of Convolutional Neural Networks (CNN) of extracting content information from an arbitrary image and style information from a well known artwork. To continue with, the style of the artwork could now be applied to an image with different content.
The key idea behind this technique is to iteratively optimise an image to match the content statistics of the arbitrary image and the style statistics of the artwork. These statistics are extracted from the images using a convolutional network. This technique is known as Neural Style Transfer.
When it comes to satellite imagery, through an artistic perspective, the two main features are land and water. Transforming a satellite image into an artistic map, using the Neural Style Transfer algorithm might not be pleasing to the eye, as the features mentioned above would not be easily distinguished. A rather more appealing result would occur with the combination of different styles applied on each feature. In other words, an artistic map would be created by "sticking" together various different pieces of stylized images, i.e. a collage. For the purposes of this task, it would be necessary to know where everything is on the image, which can be achieved by applying semantic image segmentation on the original (satellite) image. The techniques used in this study are analyzed below.

## 2. Methodology

For the purposes of this study, a Multilayer Perceptron classifier is trained for the task of semantic image segmentation and the Neural Style Transfer technique is applied for the creation of an artistic result. The utilized methods are described below.

*2.1 Semantic Segmentation*

In computer vision, image segmentation is the task in which specific regions of an image are labeled, according to what is being shown. In this study, the input is a satellite image and the output is a new image, which consists of two colors, one for each class (i.e. land or water), each pixel of which is colored according to the corresponding class. For this task, a deep learning multilayer perceptron is developed and trained, using backpropagation, which successfully distinguishes land from water, as shown in Figure 3. A multilayer perceptron (MLP) is a class of feedforward artificial neural network (ANN), which consists of an input layer, at least one hidden layer and an output layer, with two output nodes for the target classes. A standard MLP network architecture is shown in Figure 1.

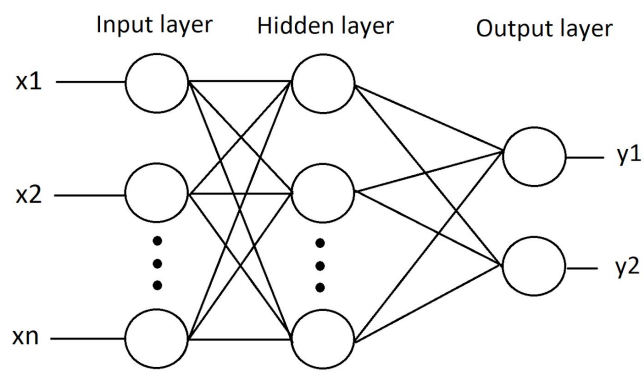

**Figure 1.** MLP architecture with an input, a hidden and an output layer.

Sentinel-2 images (geo-referenced dataset EuroSAT [3] ) were used to train the MLP classifier. This model predicts a label for each pixel and the segmented image indicates whether a region of an image consist of land or water. This information will later be used for the creation of the collage.

*2.2 Neural Style Transfer*

Neural Style Transfer is an optimization technique, which combines the content of an image (content image), e.g. a satellite image, with the style of a second image (style image), e.g. a painting of a famous artist. The output image represents the content of the content image "painted" in the style of the style image. This technique is based on Convolutional Neural Networks (CNN), and more specifically on the VGG-19 neural network architecture, which consists of 5 pooling layers for downsampling and several convolutional layers, as illustrated in Figure 2 . CNNs are a class of Deep Neural Networks that are most powerful in image processing and widely used in computer vision.They consist of layers of small computational units, which can be understood as a collection of image filters, each of which extracts a certain feature from the input image.

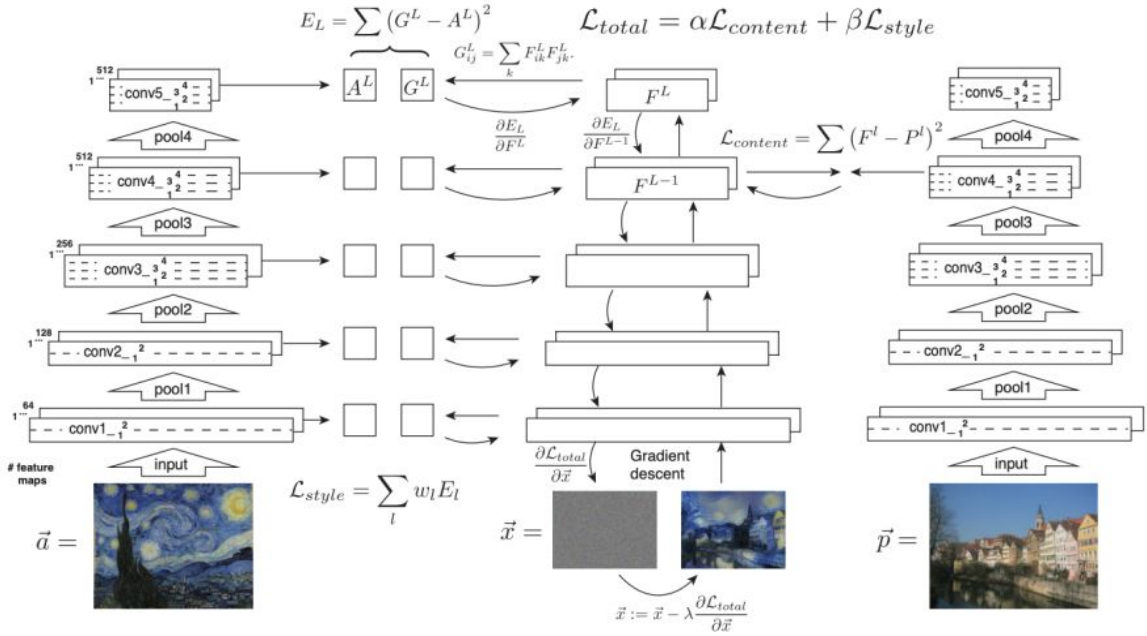

**Figure 2.** VGG-19 neural network architecture[1].

The first few layer activations of the network represent low-level features (e.g. edges and textures), while the final few layers represent higher-level features (e.g. object parts). These layers are necessary to define the representation of content and style of the images. The result is an image that combines the content of one image with the style of another,

*2.3 Collage*

At this final step, the result of the techniques described above are combined, in order to create an artistic image consisting of the two stylized images. In detail, each stylized image corresponds to a class. The values of each pixel of the stylized images are then assigned to the pixels of the segmented image of the corresponding class. The final image is a collage of the two stylized images generated at the second step, in line with the segmented image generated at the first step.

3. **Experimental Results**

The results of the method proposed above, are presented in Figure 3. The original (content) image is a Sentinel-2 image, which shows the Faroe islands. The segmented image represents the predicted class labels for each pixel in the image. Neural style transfer is then applied on the content image twice, one time for each class. The style images that were used in this case are shown in the figure below. Finally, a collage, consisting of the corresponding parts of the stylized images for each class, is created.

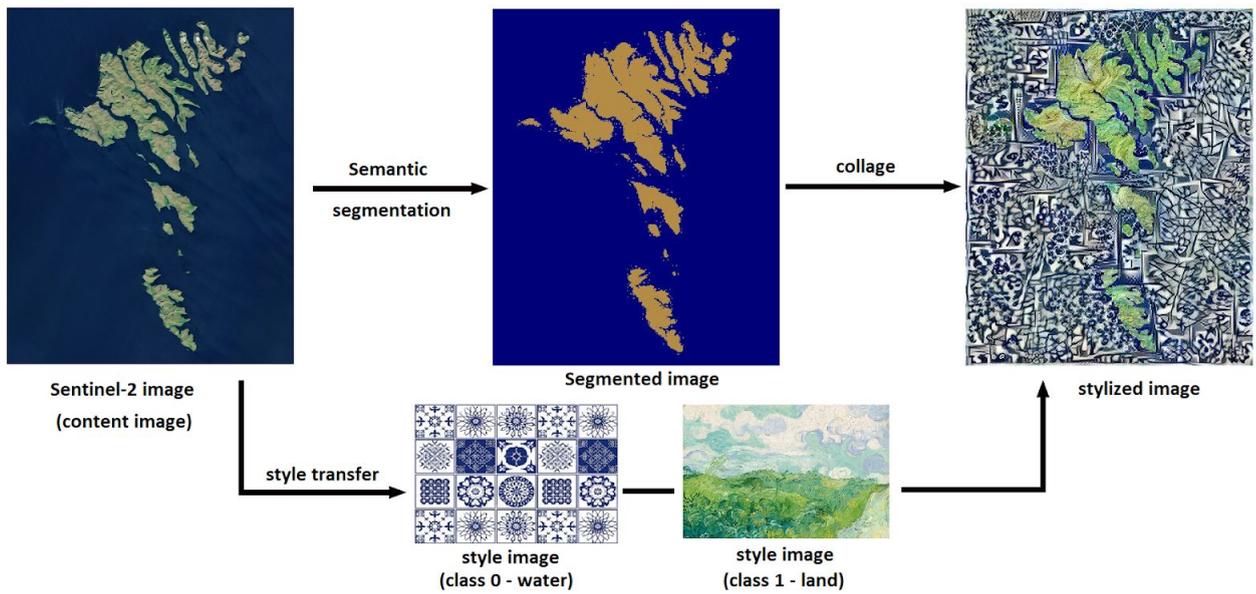

**Figure 3.** Representation of each step of the method, applied on a Sentinel-2 image of the Faroe islands.

4. **Conclusion**

As mentioned above, the aim of this study is to propose a method for creating artistic images, based on the well-known artistic technique called collage. This method can possibly be applied in the field of fashion design. Big apparel companies have already offered to the customers the experience of personal creation, by choosing colors and patterns, except now, the style of a painting can be used to stylize a design in a more artistic approach. The method described above, can also be applied in the field of gaming, for the creation of customized maps (e.g. mythical maps), allowing gamers to release their creativity.